\documentclass[letterpaper, 10 pt, conference]{ieeeconf}  % Comment this line out if you need a4paper

\IEEEoverridecommandlockouts                              % This command is only needed if 
\usepackage{amsmath}
\usepackage[hidelinks]{hyperref}
\usepackage[capitalize]{cleveref}
\usepackage{booktabs}

\usepackage{graphicx}

\usepackage{multicol}
\usepackage{multirow}
\usepackage{scalerel} %for orcid
\usepackage{tikz} % figure package
\usetikzlibrary{svg.path} %for orcid
\definecolor{orcidlogocol}{HTML}{A6CE39}
\tikzset{
  orcidlogo/.pic={
    \fill[orcidlogocol] svg{M256,128c0,70.7-57.3,128-128,128C57.3,256,0,198.7,0,128C0,57.3,57.3,0,128,0C198.7,0,256,57.3,256,128z};
    \fill[white] svg{M86.3,186.2H70.9V79.1h15.4v48.4V186.2z}
                 svg{M108.9,79.1h41.6c39.6,0,57,28.3,57,53.6c0,27.5-21.5,53.6-56.8,53.6h-41.8V79.1z M124.3,172.4h24.5c34.9,0,42.9-26.5,42.9-39.7c0-21.5-13.7-39.7-43.7-39.7h-23.7V172.4z}
                 svg{M88.7,56.8c0,5.5-4.5,10.1-10.1,10.1c-5.6,0-10.1-4.6-10.1-10.1c0-5.6,4.5-10.1,10.1-10.1C84.2,46.7,88.7,51.3,88.7,56.8z};
  }
}

\newcommand\orcidicon[1]{\href{https://orcid.org/#1}{\mbox{\scalerel*{
\begin{tikzpicture}[yscale=-1,transform shape]
\pic{orcidlogo};
\end{tikzpicture}
}{|}}}}
\title{\LARGE \bf
Mission Planning and Safety Assessment for Pipeline Inspection 
\\Using Autonomous Underwater Vehicles: A Framework based on Behavior Trees
}

\begin{document}
\author{Martin Aubard \orcidicon{0009-0000-3070-8067}, Sergio Quijano \orcidicon{0000-0002-2138-9946}, Olaya~Álvarez-Tuñón \orcidicon{0000-0003-3581-9481}, László Antal \orcidicon{0009-0005-4977-0959}, \\ Maria Costa and Yury Brodskiy \orcidicon{0009-0002-0445-8126}
% <-this % stops a space
\thanks{M. Aubard and M. Costa are with OceanScan Marine Systems $\&$ Technology, 4450-718 Matosinhos, Portugal (e-mails: \{maubard,mariacosta\}@oceanscan-mst.com). O. Álvarez is with the Aarhus University, 8000 Aarhus C, Denmark (e-mail: olaya@ece.au.dk). S. Quijano is with the IT University of Copenhagen, DK-2300 Copenhagen S, Denmark (e-mail: sequ@itu.dk), L. Antal is with RWTH Aachen University, 52074 Aachen, Germany (e-mail: antal@informatik.rwth-aachen.de) and Y. Brodskiy is with EIVA a/s, 8660 Skanderborg, Denmark (e-mail: ybr@eiva.com).
}%
}

\maketitle
\thispagestyle{empty}
\pagestyle{empty}

%%%%%%%%%%%%%%%%%%%%%%%%%%%%%%%%%%%%%%%%%%%%%%%%%%%%%%%%%%%%%%%%%%%%%%%%%%%%%%%%
\begin{abstract}
The recent advance in autonomous underwater robotics facilitates autonomous inspection tasks of offshore infrastructure. However, current inspection missions rely on predefined plans created offline, hampering the flexibility and autonomy of the inspection vehicle and the mission's success in case of unexpected events. In this work, we address these challenges by proposing a framework encompassing the modeling and verification of mission plans through Behavior Trees (BTs). This framework leverages the modularity of BTs to model onboard reactive behaviors, thus enabling autonomous plan executions, and uses BehaVerify to verify the mission's safety. Moreover, as a use case of this framework, we present a novel AI-enabled algorithm that aims for efficient, autonomous pipeline camera data collection. In a simulated environment, we demonstrate the framework's application to our proposed pipeline inspection algorithm. Our framework marks a significant step forward in the field of autonomous underwater robotics, promising to enhance the safety and success of underwater missions in practical, real-world applications. \url{https://github.com/remaro-network/pipe_inspection_mission}

\begin{keywords}
    onboard autonomy, autonomous underwater vehicle, pipeline inspection mission, behaviour trees, simulation
\end{keywords}

\end{abstract}

%%%%%%%%%%%%%%%%%%%%%%%%%%%%%%%%%%%%%%%%%%%%%%%%%%%%%%%%%%%%%%%%%%%%%%%%%%%%%%%%
\section{Introduction}
\label{sec:intro}

\begin{figure}[t]
    \centering
    \includegraphics[width=1\linewidth]{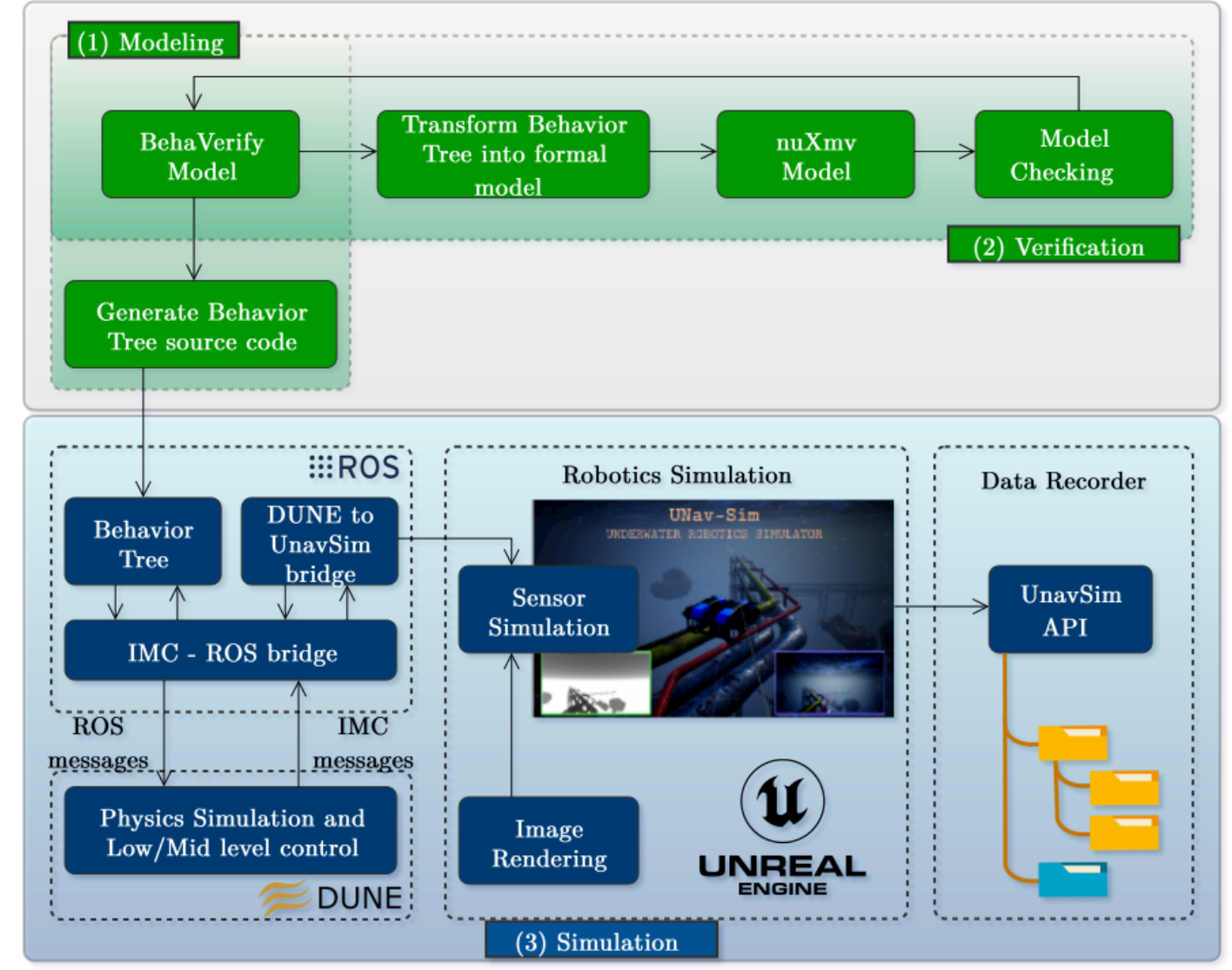}
    \caption{The proposed framework for safety assessment and behavior tree design. The green charts indicate offline design, and the blue ones indicate online execution for the mission. BehaVerify automatically generates the formal model and the behavior tree. The behavior tree is then implemented within the ROS framework to execute the mission sent to DUNE (robot's mid-level control). It retrieves its current state, fed to the UNavSim \cite{unavsim} simulator for moving the robot in the simulated environment accordingly. Meanwhile, the mission data is recorded through the UNavSim API.}
    \label{fig:GA}
\end{figure}

The increasing interest in underwater environments has led to the expansion of underwater infrastructure. Although these structures are highly beneficial, maintaining them is challenging due to the underwater environment's often unknown and complex nature. Maintenance often relies on Remotely Operated Vehicles (ROVs) manually controlled by surface-based operators for data collection \cite{CHEN201520}. However, this approach typically demands significant logistical effort and incurs high costs.

As an alternative, Autonomous Underwater Vehicles (AUVs) offer the potential for efficient, operator-independent data collection.
Present AUV-based methods for pipeline inspection, involving multiple missions for localization, data collection, and manual post-processing of pipeline data (such as camera, sonar, temperature, and salinity readings), are time-consuming. Enhancing AUV autonomy, particularly regarding onboard environmental understanding, could significantly streamline the data collection.

Methods relying on cameras \cite{rgb_pipeline_tracking} or multibeam downward-looking sonars \cite{8729803} already exist for detecting and tracking underwater pipelines. However, this technique presupposes prior knowledge of the pipeline's location, a requirement often unmet due to factors like pipeline burial. 
Instead, we propose a deep-learning-based onboard real-time pipeline detection method using Side-Scan Sonar (SSS) images. This method enables pipeline localization and following and, thus, autonomous data collection during the pipeline survey. 

However, the vehicle's increased autonomy, without operator supervision, introduces the need for new methods for maintaining safe behavior. To address this challenge, we propose the usage of Behavior Trees (BT)~\cite{Colledanchise2017} to model and verify mission plans. Behavior Trees represent mission objectives and tasks as nodes within a hierarchy, allowing for task prioritization and sequencing, ensuring the vehicle can safely and effectively adapt to dynamic scenarios. The BT design enables the use of BehaVerify~\cite{Serbinowski2022}, facilitating safety properties verification and automatic code generation for the initial implementation of the BT. We present an offline design and safety assurance framework (see~\cref{fig:GA}) for pipeline localization and inspection missions. Furthermore, this framework extends to an online underwater simulator, UNavsim, mimicking real-world scenarios, which validates the correct and safe execution of the pipeline inspection plan in a simulated environment before real-world deployment.

The paper's contribution is threefold: (1) the proposal of a novel framework for BT safety assessment, (2) the integration of an innovative onboard pipeline detection and inspection algorithm to a BT model, and (3) the design of a simulator setup for pre-deployment validation in real-world situations.

\section{Pipeline Inspection Algorithm}
\label{sec:algorithm}
\subsection{Algorithm}
We propose an algorithm that employs an object detection deep learning (DL) model using Side-Scan Sonar images to detect and localize the pipeline in real-time, using the Light Autonomous Underwater Vehicle (LAUV) \cite{SOUSA2012268}. This algorithm allows tracking the pipeline until it has been fully inspected from beginning to end. Additionally, the pipelines might be buried beneath or lying over the sand; therefore, the algorithm incorporates a search maneuver to relocate the pipeline if it remains undetected for a specified duration. 

The algorithm switches between three core maneuvers: \textit{Rows}, \textit{GoTo}, and \textit{Tracking}. The mission starts with the \textit{Rows} maneuver, which allows the vehicle to survey a large region using the SSS to detect the pipeline. Upon successful detection, the vehicle switches to a \textit{GoTo} maneuver, steering itself toward the identified coordinates. Once near the pipeline (with the distance provided by the DL model), the vehicle's maneuver changes to \textit{Tracking}, consistently maintaining an optimal distance to track the pipeline. If the DL model does not detect the pipeline for more than a certain amount of time, we assume that the pipeline is not detectable anymore. The vehicle then reverts to the \textit{Rows} maneuver, entering search mode and activating a timer. It continues searching until the pipeline is detected again or the timer reaches its limit. If the pipeline reappears during the search, the algorithm repeats the previously described steps until the pipeline is lost once more; however, if the timer is up while in the search mode, the vehicle initiates a \textit{GoTo} maneuver towards the last known pipeline location. 

\begin{figure*}[ht!]
    \centering
    % \huge
    \includegraphics[width=0.9\textwidth]{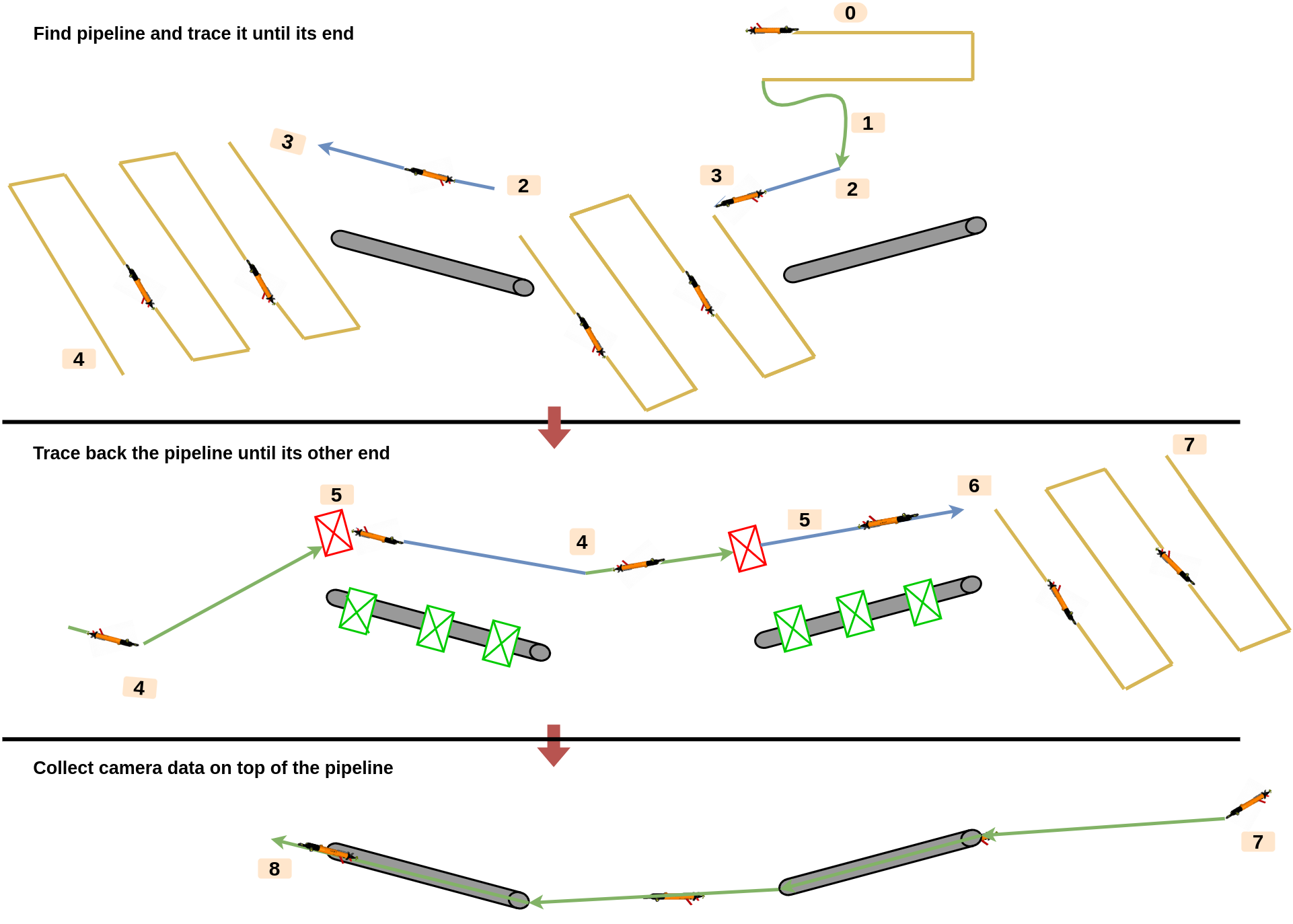}
    \caption{Pipeline Tracking. This scheme depicts two pipelines, represented in grey, which could also illustrate a single pipeline partially obscured by sand. The orange squares indicate the algorithm step. The diagram characterizes the three different maneuvers, each marked by arrows in distinct colors: the \textit{Rows} maneuver in yellow, the \textit{GoTo} maneuver in green, and the \textit{Tracking} maneuver in blue. The green squares mark the pipeline's detected and stored latitude/longitude positions. Finally, the red squares represent the AUV's position at the last pipeline detection.}
    \label{fig:pipeline_tracking}
\end{figure*}

Considering the vehicle is at the pipeline's beginning, the objective is to track back and record every pipeline detected position (latitude and longitude) until the pipeline's end. This process continues until the second timer is up, indicating that the pipeline has been tracked from beginning to end. Then, the vehicle switches to a \textit{GoTo} maneuver covering all identified segments. 
This strategy positions the vehicle directly over the pipeline for easier RGB data acquisition. The algorithm's steps are represented in \cref{fig:pipeline_tracking}

% \begin{table*}[htbp]
% %\begin{wraptable}{r}{0.5\textwidth}
% \centering
% % \captionsetup{font=large}
% \caption{IMC Messages Description}
% \label{IMC Messages Description}
% \renewcommand{\arraystretch}{0.95} 
% %\resizebox{\textwidth}{!}{
% \scalebox{1.3}{
% \small
% \begin{tabular}{l|c}
% %\hline
% \textbf{IMC Messages} & \textbf{Parameters} \\
% \hline
% CVDL Output & targetID, confidence, side, x, y \\
% \hline
% Detected Target State & targetID, confidence, targetLong, targetLat, side, distance, targetAngle \\
% \hline
% Tracking-State & desiredDistance, expectedHeading, targetLat, targetLong \\
% \hline
% \end{tabular}
% }
% %\end{wraptable}{r}{0.5\textwidth}
% \end{table*}

% \begin{table*}
%     \centering
%     \resizebox{\textwidth}{!}{%
%     \begin{tabular}{p{0.3\textwidth} p{0.45\textwidth}}
%         % \toprule
%         IMC Messages & Parameters\\
%         \midrule
%         CVDL Output & targetID, confidence, side, x, y\\
%         %-----------------------------------------
%          &  \\
%          Detected Target State & targetID, confidence, targetLong, targetLat, side, distance, targetAngle\\
%         %-----------------------------------------
%          &  \\ 
%          Tracking-State & desiredDistance, expectedHeading, targetLat, targetLong\\
%         %-----------------------------------------
%         \bottomrule
%     \end{tabular}%
%     }
%     \caption{IMC Messages Description.}
%     \label{IMC Messages Description}
% \end{table*}

\subsection{Real-time pipeline detection}
The proposed pipeline tracking algorithm is based on real-time pipeline detection using the YOLOX object detection model \cite{yoloxpaper}, trained on the SubPipe dataset \cite{subpipe_dataset}. SubPipe offers a diverse range of pipeline data, including RGB, Grayscale, and SSS images. As detailed in the algorithm section, for effective \textit{GoTo} and \textit{Tracking} maneuvers, the object detection model must provide information about the pipeline, including its classification and the distance of the AUV from the detected pipeline.
Aubard et al. \cite{rw:sonarwalldet} have previously demonstrated real-time distance measurement to a wall, and our paper adopts a similar approach but focuses on pipeline data. They assumed that the predicted bounding box's middle point is the object's position and estimated the distance from the vehicle by considering the SSS configuration parameters, such as frequency and range affecting SSS raw data resolution and the AUV's heading.

Our method extends this principle by converting the current distance into the latitude/longitude position of the detected pipeline. SSS data inherently involves a time lag as it collects past data. Consequently, the detected distance reflects a past rather than a current position, introducing uncertainties for onboard guidance. This guidance principle relies on the pipeline's two most recent latitude/longitude detections, using them to form a projected line. The AUV follows this line based on the latitude/longitude position rather than using outdated distance data. Those coordinates are calculated based on the estimated state of AUV at the detection timestamp.
% To implement this approach, onboard detection information, which provides the detected distance, must be converted to the latitude/longitude position of the pipeline. The LAUV, equipped with Dune as its onboard software and an Inertial Navigation System (INS), calculates the estimated position of the vehicle based on the last GPS communication (at the surface) and its previous motion. Therefore, processing the distance, the heading, and the AUV's position during pipeline detection is essential to transform the detected pipeline distance into a corresponding latitude/longitude position.
\\

% \begin{equation}
%     % \begin{aligned}
%     \text{Lat}_{\text{pipeline}} &= \text{Lat}_{\text{AUV}} + \Delta d \cdot \cos(\theta)
%     % \end{aligned}
% \end{equation}

% \begin{equation}
%     % \begin{aligned}
%     \text{Long}_{\text{pipeline}} &= \text{Long}_{\text{AUV}} + \Delta d \cdot \sin(\theta)
%     % \end{aligned}
% \end{equation}

% In this equation:
% \begin{itemize}
%     \item $\text{Lat}_{\text{pipeline}}$ and $\text{Long}_{\text{pipeline}}$ are the latitude and longitude of the detected pipeline, respectively.
%     \item $\text{Lat}_{\text{AUV}}$ and $\text{Long}_{\text{AUV}}$ are the latitude and longitude of the AUV at the time of pipeline detection.
%     \item $\Delta d$ is the detected distance to the pipeline from the AUV.
%     \item $\theta$ is the heading of the AUV.
% \end{itemize}
% All those values are at the same timestamp as the SSS raw data of the detected pipeline, given by the bounding box's middle point).

Furthermore, to enable interaction between the detected pipeline and the LAUV, DUNE (onboard software) utilizes its onboard DL version \cite{LSTSDL}. It facilitates onboard data processing and establishes new Inter-Module Communication (IMC) messages \cite{LSTS-Toolchain} between the DL model and DUNE, following the LSTS toolchain requirements. 
% The three messages utilized for the pipeline detection and tracking use case are \textit{CVDL Output}, \textit{Detected Target State}, and \textit{Tracking State}. Their definitions are listed in Table 1.

% This strategy ensures the vehicle is precisely over the pipeline, facilitating the acquisition of RGB data. 

\section{Framework for Modeling and Verification of Behavior Trees}
\label{sec:bt_ver}
%Behavior trees are hierarchical control structures that orchestrate autonomous systems' decision-making processes. They consist of control nodes (e.g., selectors and sequences) that define decision flow and action nodes representing specific behaviors~\cite{Colledanchise2017}. BTs enable representing mission objectives and tasks as nodes within a hierarchy, allowing for task prioritization and sequencing, ensuring the vehicle can adapt to dynamic scenarios effectively. Monitors and checks continuously assess the system's state, ensuring it adheres to safety constraints throughout its mission.

%Behavior trees consist of control nodes (e.g., selectors and sequences) that define decision flow and action nodes representing specific behaviors~\cite{Colledanchise2017}, which makes them suitable for mission planning and safety assessment in our pipeline inspection algorithm.
Our framework consists of three phases: \textit{Modeling}, \textit{Verification}, and \textit{Simulation}, as shown in~\cref{fig:GA}. It leverages BehaVerify~\cite{Serbinowski2022} to build an abstract model that can be used for verification and automatic code generation of the behavior tree. In the following, we illustrate the application of our framework to the proposed pipeline inspection algorithm.

\subsection{Modeling}
A behavior tree is a widely used control structure in artificial intelligence and robotics for modeling the decision-making process of autonomous agents. It consists of a hierarchical structure of nodes representing actions, conditions, and control flow elements~\cite{Colledanchise2017}. These nodes can be categorized into three main types: control flow nodes, task nodes, and condition nodes. 

Control flow nodes determine the structure and flow of the tree, including sequences, selectors, and decorators. Task nodes represent the agent's atomic actions, such as moving to a location or following a target. Condition nodes evaluate certain conditions, such as the presence of a pipeline segment or the agent's status. The tree is traversed recursively from the root node down to the leaf nodes, with each node determining the next course of action based on its logic. Control flow nodes determine the order in which child nodes are executed, while task and condition nodes perform specific actions or evaluate conditions~\cite{Iovino2022}.

\begin{figure}[ht!]
    \centering
    \includegraphics[width=.47\textwidth]{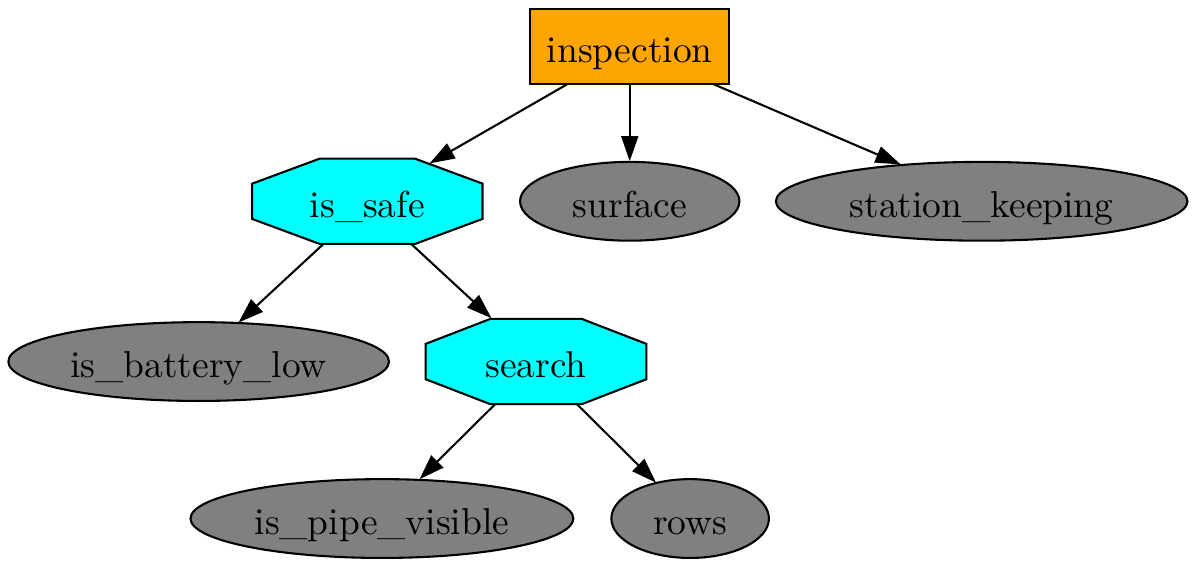}
    \caption{Fragment of pipeline inspection's Behavior Tree. Cyan nodes are selector nodes; the left grey nodes on each selector subtree are check nodes, and the right nodes are action nodes. The orange node is a sequence node.}
    \label{fig:inspection_bt}
\end{figure}

For instance,~\cref{fig:inspection_bt} depicts a fragment of our BT model. It models a scenario when the pipeline is not detected within a set time. The AUV should decide if a new search phase can be executed while also considering its power autonomy. We use a selector flow with a check node that reads the DL detection output; if no detection is recorded, a task node commands the AUV to initiate the search maneuver. For the battery check, we use a selector flow, in which a check node reading the vehicle's battery status decides if a task node commanding the AUV to abort the mission, resurface, and station-keep should be executed or it should proceed with the mission.
%Behavior Trees are suitable for mission planning and safety assessment in our pipeline inspection algorithm. For instance, when the pipeline is not detected within a set time, the AUV should revert to the rows maneuver in search of the pipeline again. In our mission BT, we model this scenario as a selector flow, with a check node that reads the deep learning detection output; if no detection is recorded, then an action node commands the AUV to initiate the Rows maneuver. At the same time, we want to guarantee that the AUV power supply is enough to execute this behavior. Therefore, we model a battery check scenario as a selector flow, in which a check node reading the vehicle's battery status decides if a node action commanding the AUV to abort the mission, resurface, and station-keep in place should be executed or the AUV should continue with the rows maneuver.

\begin{figure*}[!htbp]
    \centering
    \includegraphics[width=\textwidth]{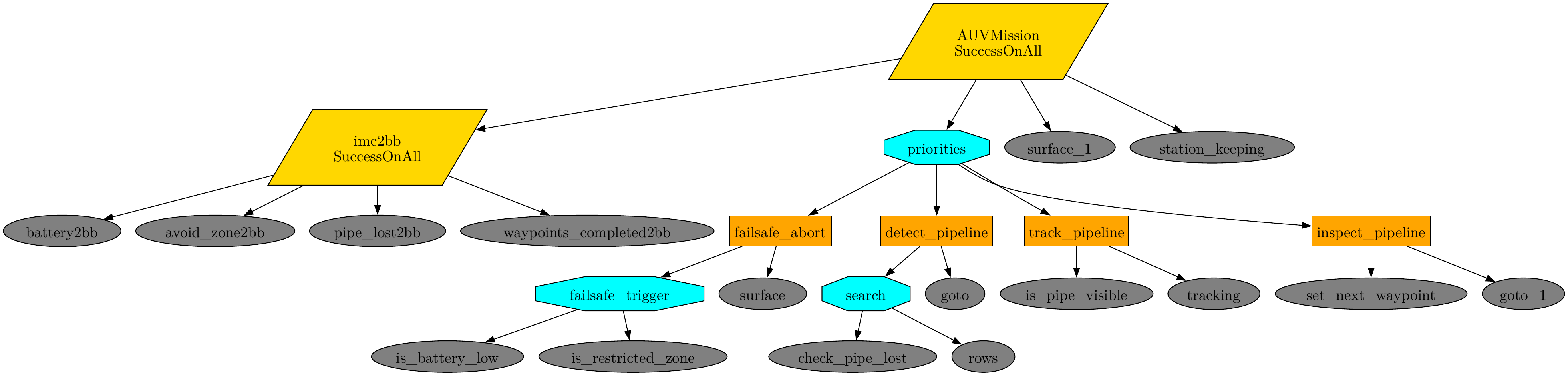}
    \caption{Pipeline inspection's Behavior Tree. In addition to the functional requirements for a pipeline inspection mission, the BT includes safety checks and subscription tasks to ROS/IMC messages for later use in simulation. The different phases of our algorithm are organized under the \textit{priorities} subtree; the \textit{failsafe\_trigger} sequence checks if a battery low warning or restricted zone warning is raised; in such a case, it executes an emergency surface maneuver, and stops the mission. The \textit{detect\_pipeline} sequence executes the search phase if the pipeline is not detected; otherwise, it executes a goto maneuver to approach the pipeline and start with the next phase. With the \textit{track\_pipeline} sequence, the AUV navigates, as long as the pipeline is detected, along the pipeline and stores the position where a pipeline segment is detected. Finally, with the \textit{inspect\_pipeline} sequence, we command the AUV to navigate all the identified pipeline segments for RGB data acquisition. The mission ends with the AUV going to the surface and executing a station-keep maneuver.}
    \label{fig:auvmission_bt}
\end{figure*}

Considering the functional aspects described in~\cref{sec:algorithm}, the complete BT model is depicted in ~\cref{fig:auvmission_bt}. The BT model includes the definition of tasks that subscribe to IMC/ROS messages and stores the needed variable values in the BT blackboard. Variables indicate the pipeline's detection (through the DL algorithm), low-level battery warnings, and the completion of navigation along the discovered locations. These tasks are grouped under a subtree using a \textit{Parallel} node that returns a successful status only when all the subnodes return success. We encode these nodes into a BehaVerify abstract model. 

Taking advantage of the order of execution provided by a selector node, the BT executes the different phases in the order prescribed by our algorithm. In addition to the search, tracking, and inspection functional requirements, under the \textit{priorities} selector node, we preempt the validation of a safe operation of the AUV by adding a fail-safe sequence in case the vehicle's battery is low or when the vehicle is approaching a restricted zone. The BT begins executing when it receives and propagates a tick event throughout the tree. The leaf nodes execute checks on indicators of interest (e.g., the battery level or the distance to the pipeline) or actions, such as a Goto maneuver. Standard \textit{Running}, \textit{Failure}, and \textit{Success} return types and node semantics are applied in the implementation of each node of our BT to execute the different phases of our algorithm appropriately.

Using the Domain Specific Language provided by BehaVerify and their corresponding transformation tools, with a single input abstract model, we can generate the initial implementation of the BT automatically.
%Ensuring the safety of autonomous vehicles is a top priority in their development and deployment. Behavior trees allow to define explicitly safety-critical behaviors and conditions within the control structure. These behaviors encompass actions that must be executed to prevent accidents or undesirable outcomes.

\subsection{Verification}
In addition to the operational and safety strategies, within the same abstract model, we define safety specifications as Linear Temporal Logic (LTL)~\cite{Pnueli1977} properties that can be analyzed with model-checking techniques~\cite{Baier2008}; specifically, the abstract model is translated to a formal model in the nuXmv model checker format~\cite{Cavada2014} and can be iteratively used to verify the specified safety properties and adapt the BT according to the obtained results. In our battery check scenario, we specify the property $G((batteryLow=1) \Rightarrow (stationKeepingTask.active))$ to guarantee that the AUV remains in station-keeping if the battery level is not enough to continue with the mission. 

Furthermore, we want to ensure that the mission is executed as expected. For instance, if the station-keeping task is inactive, we want to ensure that the AUV executes the search, tracking, or inspection actions according to our algorithm; LTL properties \ref{prop_search}, \ref{prop_track}, and \ref{prop_inspect}, respectively, are specified to verify that there is adequate progress in the AUV's mission:
%\newline

%\begin{align*}
%& G(\neg(missionStatus = m\_station \lor \nonumber \\
%& missionStatus = m\_surface) \Rightarrow \nonumber \\ 
%& ((RowsTask.active \land pipeLost=1 \land \nonumber \\
%& missionStatus = m\_search) \lor \nonumber \\
%& (TrackingTask.active \land \nonumber \\
%& pipeLost=0 \land missionStatus = m\_tracking) \lor \nonumber \\
%& (GotoTask.active \land missionStatus = m\_steering))) \nonumber
%\end{align*}
\begin{equation}\label{prop_search}
\begin{aligned}
& G(\neg (status = m\_station \lor status = m\_surface) \Rightarrow \\ 
& (RowsTask.active \land pipeLost=1 \land \\
& status = m\_search)) 
\end{aligned}
\end{equation}

\begin{equation}\label{prop_track}
\begin{aligned}
& G(\neg (status = m\_station \lor status = m\_surface) \Rightarrow \\ 
& (TrackingTask.active \land \\
& pipeLost=0 \land status = m\_tracking))
\end{aligned}
\end{equation}

\begin{equation}\label{prop_inspect}
\begin{aligned}
& G(\neg (status = m\_station \lor status = m\_surface) \Rightarrow \\ 
& (GotoTask.active \land status = m\_steering))
\end{aligned}
\end{equation}

%\newline

If the properties are violated, the model checker returns a counterexample that can be used to manually refine the BT to ensure that the safety/functional scenario is executed as expected. The same abstract model used in the previous phase is used to specify the LTL properties of interest and translated to the nuXmv automatically to execute the verification using the existing model checker tools.

\subsection{Simulation}
The BT is integrated within a ROS-based simulation framework. The DUNE simulator is the runtime environment for vehicle onboard software that implements the simulation of the LAUV's mathematical model and allows the control of the simulated vehicle through IMC messages. 
The UNavSim simulator carries out the sensor simulation, including realistic renderings from the camera and segmentation annotations \cite{unavsim}. UNavSim and DUNE are synchronized through a ROS bridge that handles the communication between the two.
This approach allows us to verify the correct execution of the pipeline inspection plan in a simulated environment before proceeding to real-world deployment. It also serves as a safety control mechanism, strengthening the algorithm's readiness for practical implementation.

\begin{table*}
    \centering
    \resizebox{\textwidth}{!}{%
    \begin{tabular}{lllll}
        % \toprule
        Package & Publisher node & Topic & Type & content \\
        \midrule
        \texttt{unavsim\_ros\_pkgs} & \texttt{unavsim\_node} & \texttt{/<camera\_name>/Scene} & \texttt{sensor\_msgs/msg/Image} & UNavSim RGB camera\\
        %-----------------------------------------
         &  & \texttt{/<camera\_name>/Segmentation} & \texttt{sensor\_msgs/msg/Image} & UNavSim segmentation labels\\
        %-----------------------------------------
         &  & \texttt{/<camera\_name>/DepthPlanar} & \texttt{sensor\_msgs/msg/Image} & UNavSim depth camera\\
        %-----------------------------------------
         &  & \texttt{/<camera\_name>/Scene/camera\_info} & \texttt{sensor\_msgs/msg/CameraInfo} & UNavSim camera intrinsics\\
        %-----------------------------------------
         &  & \texttt{/imu/Imu} & \texttt{sensor\_msgs/msg/Imu} & UNavSim's IMU measurement\\
        %-----------------------------------------
         & & \texttt{/altimeter/barometer} & \texttt{unavsim\_interfaces/msg/Altimeter} & UNavSim altimeter measurements\\
        %-----------------------------------------
         & & \texttt{/tf} & \texttt{tf2\_msgs/msg/TFMessage} & 6 DOF pose in UNavSim\\
        %-----------------------------------------
        \texttt{imcpy\_ros\_bridge} & \texttt{imc2ros} & \texttt{/from\_imc/base\_link} & \texttt{geometry\_msgs/msg/PoseStamped} & 6 DOF pose in DUNE \\ 
        %-----------------------------------------
         &  & \texttt{/from\_imc/estimated\_state} & \texttt{imc\_ros\_msgs/msg/EstimatedState} & 6 DOF pose estimated by DUNE\\
        \bottomrule
    \end{tabular}
    }
    \caption{Data published by the ROS framework with the UNavSim simulation.}
    \label{tab:ros_msgs}
\end{table*}

\begin{figure}
    \centering
    \includegraphics[width=0.49\linewidth]{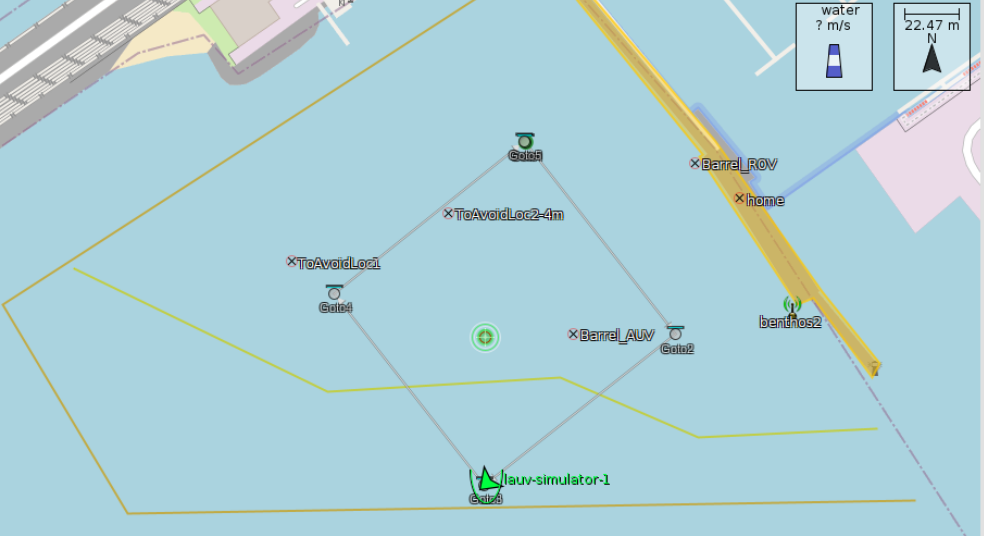}
    \includegraphics[width=0.49\linewidth]{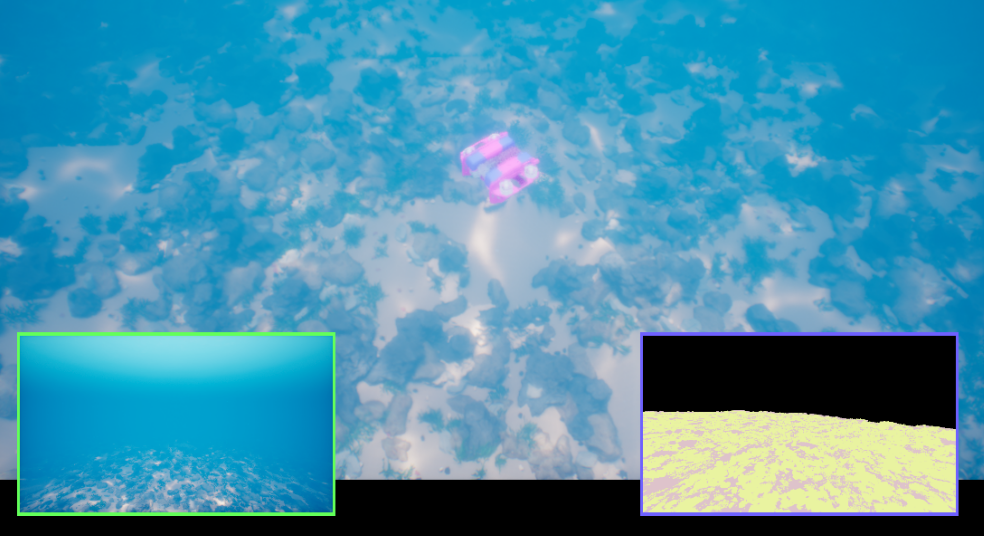}
    \caption{The integrated simulation framework allows synchronization between DUNE and UNavSim. On the left is the Neptus interface, the primary control and monitoring tool for the DUNE system. On the right is the UnavSim simulator, displaying realistic renderings of the environment and the underwater robotic vehicle. The UnavSim interface displays the simulation of the RGB camera and the corresponding segmentation labels.}
    \label{fig:neptus-unavsim-screenshots}
\end{figure}

The ROS simulation framework is structured as two ROS stacks separated by functionality: \texttt{remaro\_uw\_sim} and \texttt{imcpy\_ros\_bridge}. 
The \texttt{remaro\_uw\_sim} stack serves as an integrative bridge, efficiently linking the UNavSim simulation with DUNE and ROS. Within this stack, the \texttt{unavsim\_ros\_pkgs} package utilizes the \texttt{unavsim\_node} to interface directly with UNavSim, acquiring sensor measurements and effectively publishing these as ROS topics. Additionally, \texttt{mimir\_node}, building upon the methodologies presented in \cite{alvarez2023mimir}, integrates a dataset recorder. This recorder automatically captures sensor data via the UNavSim API.
Furthermore, the \texttt{neptusROV} package within the same stack synchronizes the robot's pose between the DUNE and UNavSim environments. It achieves this by reading the robot's pose data, as communicated by DUNE via IMC messages, and correspondingly adjusting the UNavSim's robot model to mirror this pose. 

On the other hand, the \texttt{imcpy\_ros\_bridge} stack interfaces between DUNE, ROS, and the behavior trees.
Central to this stack is the \texttt{imc\_ros\_msgs} package. It contains custom ROS message definitions, allowing the publishing and subscribing of IMC messages through ROS topics. 
The \texttt{imcpy\_ros\_bridge} package contains the \texttt{imc2ros} node that establishes a connection with DUNE and allows a bidirectional data exchange through the custom ROS messages defined in \texttt{imc\_ros\_msgs}.
Finally, the \texttt{imcpy\_trees} package includes exemplary implementations of \texttt{py\_trees} behavior trees to control the robot using the \texttt{imcpy\_ros\_bridge}.

\begin{figure}
    \centering
    \includegraphics[width=0.32\linewidth]{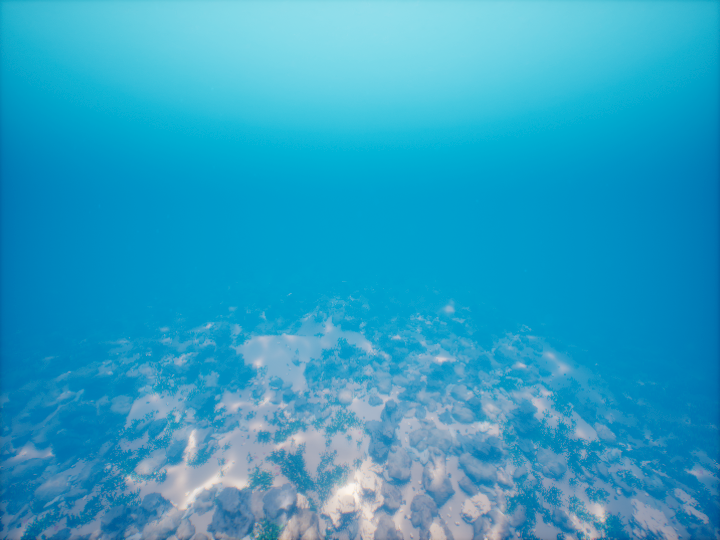}
    \includegraphics[width=0.32\linewidth]{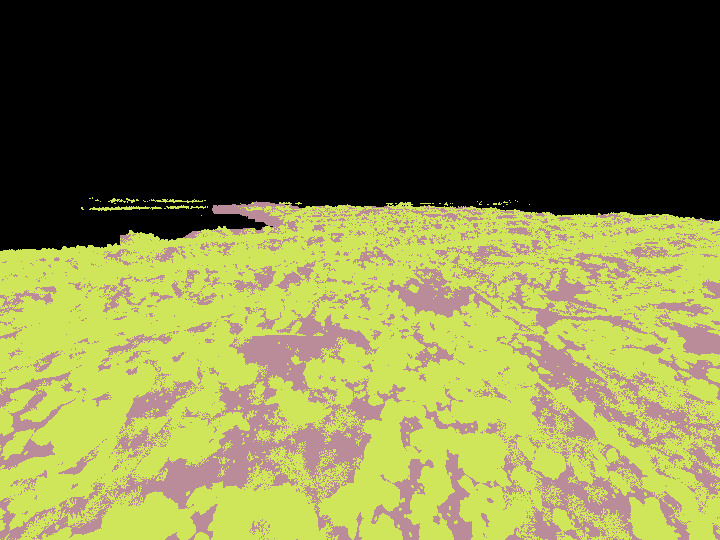}
    \includegraphics[width=0.32\linewidth]{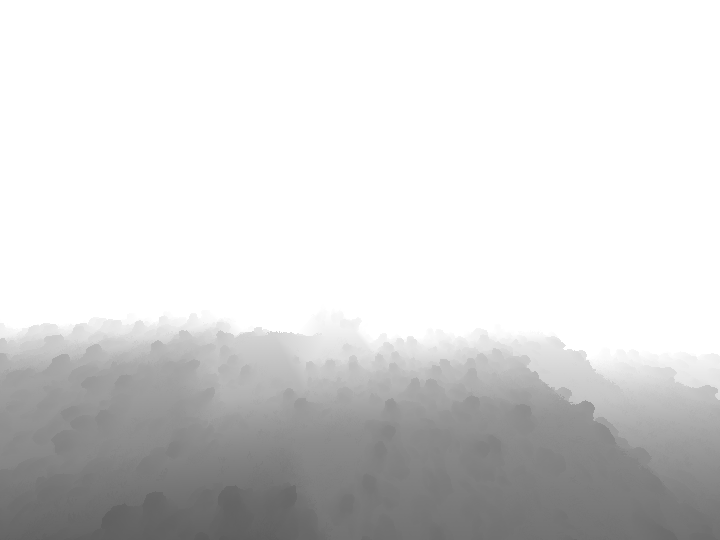}
    \caption{Example of the RGB (left) segmentation (center) and depth (right) images rendered by the UNavSim simulator and recorded with the proposed framework.}
    \label{fig:imgs-example}
\end{figure}

\begin{figure}
    \centering
    \includegraphics[width=\linewidth]{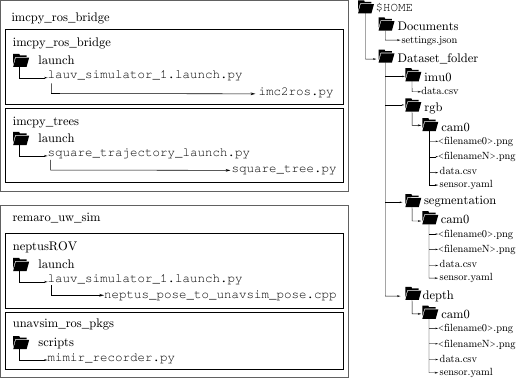}
    \caption{Framework for the dataset recording of the Leixões marina example. On the left are the ROS stacks and packages the framework requires, highlighting the specific launch files employed and the nodes they activate. The key files involved in the dataset recording process are depicted on the right.  UNavSim requires the \textit{settings.json} file to configure the simulation. The \textit{Dataset folder }shows an example of the dataset structure generated by the recording script.}
    \label{fig:example_simulation_framework}
\end{figure}

The following section describes how the proposed framework is used for an example of data recording in a simulated environment modeled after the Leixões marina:

\subsubsection{The Leixões marina example}
The geospatial alignment between the DUNE framework and the UNavSim simulation is achieved through the Cesium plugin \cite{cesium}. This alignment process involves setting a common geographical origin in both systems, ensuring that the starting point of the robotic mission is identical in DUNE and UNavSim.
The UNavSim environment includes a seabed mesh generated from the bathymetric map of Leixões. It is covered with rocks and vegetation, thereby replicating the actual conditions of the seabed with high fidelity.  Furthermore, the simulation includes light reflections and scattering effects (see Fig. \ref{fig:neptus-unavsim-screenshots}).

An overview of the simulation and data recording framework is depicted in Fig. \ref{fig:example_simulation_framework}. 
The dataset recording process is executed as follows: the launch files \texttt{lauv\_simulator\_1.launch.py} in the \texttt{imcpy\_ros\_bridge} and the \texttt{neptusROV} packages handle the connection between DUNE, UNavSim and ROS. Moreover, the mirroring of the robot's pose from DUNE to UNavSim is carried out by the \texttt{neptus\_pose\_to\_unavsim\_pose.cpp} executable. The \texttt{square\_tree.py} node includes a simple example of a \texttt{py\_tree} using \texttt{IMC} messages to make the robot draw a square trajectory. Running the script \texttt{mimir\_recorder.py} simultaneously with the mission records the sensor data into the filesystem following the dataset structure exemplified in Fig. \ref{fig:example_simulation_framework}. A sample of the images rendered by the RGB, segmentation, and depth cameras is shown in Fig. \ref{fig:imgs-example}. Alternatively, given the ROS topics showcased in Table \ref{tab:ros_msgs}, the dataset can also be recorded as a \textit{rosbag}.

%\section{Simulation}
%\label{sec:simulation}
%\input{tex/04_simulation}

\section{Conclusion and Future Work}
\label{sec:conclusion}
We propose a comprehensive framework that utilizes Behavior Trees (BT) as a powerful tool for modeling. This framework not only facilitates the seamless integration of safety scenarios into mission planning and execution but also leverages robust offline verification techniques. In a simulated environment, we demonstrate the application of this framework to an innovative pipeline detection and inspection algorithm.
This framework enhances offline safety verification for underwater missions. It models these missions using a BT architecture and employs model-checking techniques for initial verification, followed by simulation of the onboard BT behavior in an underwater simulator. This approach is pivotal in understanding and improving the safety and reliability of underwater missions.

However, challenges remain in defining comprehensive safety properties that account for all potential unplanned sensor malfunctions and behavioral anomalies during field maneuvers. Additionally, the current limitations of the UNavSim simulator in simulating realistic side-scan sonar images pose a significant challenge, introducing uncertainties in verifying the effectiveness of detection behaviors. Addressing these challenges will further enhance the framework's efficacy in real-world deployments.

%This modeling approach significantly enhances offline verification processes before field deployment. By rigorously testing and validating behaviors in a controlled simulation environment, the likelihood of successful and safe mission execution in real-world scenarios is significantly increased.
%Future work will focus on: i. extending the simulator to encompass the pipeline detection and tracking use case; ii. validating the safety properties and the BT in UNavSim and actual field conditions.
Moving forward, our future efforts will focus on expanding the simulator to encompass pipeline detection and tracking scenarios, validating safety properties and BT performance in both simulated and actual field conditions. By continually refining and validating our framework, we aim to advance the safety and efficiency of autonomous underwater missions, ultimately contributing to sustainable offshore infrastructure management.

\section*{Acknowledgements}
\noindent This project has received funding from the European Union's EU Framework Programme for Research and Innovation Horizon 2020 under the Grant Agreement No 956200. For further information, please visit \url{https://remaro.eu}.

% This project has received funding from the European Union's Horizon 2020 research and innovation programme under the Marie Skłodowska-Curie grant agreement No. 956200. For further information, please visit \url{https://remaro.eu}.

\addtolength{\textheight}{-12cm}   % This command serves to balance the column lengths

%%%%%%%%%%%%%%%%%%%%%%%%%%%%%%%%%%%%%%%%%%%%%%%%%%%%%%%%%%%%%%%%%%%%%%%%%%%%%%%%

%%%%%%%%%%%%%%%%%%%%%%%%%%%%%%%%%%%%%%%%%%%%%%%%%%%%%%%%%%%%%%%%%%%%%%%%%%%%%%%%

%%%%%%%%%%%%%%%%%%%%%%%%%%%%%%%%%%%%%%%%%%%%%%%%%%%%%%%%%%%%%%%%%%%%%%%%%%%%%%%%

% \section*{Acknowledgment}

% This project has received funding from the European Union’s EU Framework Programme for Research and Innovation Horizon 2020 under the Grant Agreement No 956200.

\bibliographystyle{ieeetr}
\bibliography{literature.bib}

\end{document}